\documentclass{article}

\usepackage{spconf,amsmath,graphicx}
\usepackage{enumitem}
\usepackage{graphicx}
\usepackage{amsfonts}
\usepackage{algorithm}
\usepackage{algorithmic}
\usepackage{booktabs}
\usepackage{multirow}

\usepackage{float}
\usepackage[letterpaper, top=1in, bottom=1.1in, left=0.62in, right=0.625in]{geometry}

\title{Achieving Fine-grained Cross-modal Understanding through Brain-inspired Hierarchical Representation Learning}
\name{Weihang You$^{*}$, Hanqi Jiang$^{*}$, Yi Pan, Junhao Chen, Tianming Liu, Fei Dou$^{\dagger}$
\thanks{$^*$Equal Contribution.}
\thanks{$\dagger$Corresponding Author.}
}
\address{School of Computing, University of Georgia, Athens, GA, USA}
\begin{document}
%\ninept
%
\maketitle
\begin{abstract}
Understanding neural responses to visual stimuli remains challenging due to the inherent complexity of brain representations and the modality gap between neural data and visual inputs. Existing methods, mainly based on reducing neural decoding to generation tasks or simple correlations, fail to reflect the hierarchical and temporal processes of visual processing in the brain. To address these limitations, we present NeuroAlign, a novel framework for fine-grained fMRI-video alignment inspired by the hierarchical organization of the human visual system. Our framework implements a two-stage mechanism that mirrors biological visual pathways: global semantic understanding through Neural-Temporal Contrastive Learning (NTCL) and fine-grained pattern matching through enhanced vector quantization. NTCL explicitly models temporal dynamics through bidirectional prediction between modalities, while our DynaSyncMM-EMA approach enables dynamic multi-modal fusion with adaptive weighting. Experiments demonstrate that NeuroAlign significantly outperforms existing methods in cross-modal retrieval tasks, establishing a new paradigm for understanding visual cognitive mechanisms.
\end{abstract}
\section{Introduction}
\label{sec:intro}

Understanding how the brain encodes visual information is a fundamental challenge in neuroscience. Recent advances in fMRI acquisition and deep learning have enabled new approaches to decode visual content from brain activity \cite{horikawa2017generic,huth2016natural,chen2023seeing,mindvis}. However, achieving fine-grained alignment between fMRI signals and corresponding video frames remains difficult.

The challenge stems from fundamental properties of fMRI data. Brain responses are temporally delayed and smeared by the hemodynamic response function (HRF), exhibit low signal-to-noise ratio due to physiological artifacts, and reflect hierarchical processing across multiple cortical areas. These characteristics make precise temporal and spatial alignment with visual stimuli inherently difficult.

Existing methods fall into two categories: correlation-based mapping \cite{horikawa2017generic,huth2016natural,jiang2012fmri,lohmann2014correlation} and generation-based approaches using vision-language models \cite{chen2023seeing,gong2025neuroclips,mindvis,huang2023functional,zheng2024llm4brain}. CLIP-style contrastive learning \cite{clip,yao2021filip,jia2021scaling,xu2021videoclip,xue2022clip} has shown strong performance in vision-language tasks and inspired neural decoding applications \cite{chen2023cinematic,scotti2023reconstructing,scotti2024mindeye2}. However, continuous embeddings from CLIP-based methods are brittle under HRF variability, noise, and hierarchical coding. Fine-grained correspondences become unstable.

We address these limitations through discrete representations. Vector quantization \cite{van2017neural} provides noise-robust, temporally stable codes that naturally accommodate hierarchical matching. Recent advances in video discretization and spatiotemporal self-supervision \cite{guo2024crossmae,fu2024linguistic,preechakul2022diffusion,he2024anomalycontrol,qian2021spatiotemporal,dave2022tclr,misra2024towards} demonstrate the effectiveness of discrete representations for complex visual data.

In this paper, we present NeuroAlign, a framework for fine-grained fMRI multi-modal alignment inspired by the hierarchical organization of the human visual system \cite{konen2008two}. Our approach integrates three components: Neural-Temporal Contrastive Learning (NTCL) for global semantic alignment with temporal modeling, VQ-based pattern matching for fine-grained correspondences, and DynaSyncMM-EMA for balanced multi-modal fusion. On the CC2017 dataset, NeuroAlign achieves significant improvements over state-of-the-art methods across all cross-modal retrieval tasks.
\\
Our contributions are:
\begin{itemize}
\item We propose NeuroAlign, a discrete representation learning framework that addresses the core challenges of fMRI-video alignment: temporal mismatch, low SNR, and multi-scale hierarchical processing.
\item We introduce Neural-Temporal Contrastive Learning (NTCL) that explicitly models HRF-induced delays and temporal dependencies for stable cross-modal semantic alignment.
\item We develop DynaSyncMM-EMA, a synchronized codebook update mechanism with variance-aware weighting that balances multi-modal contributions without erasing modality-specific structure.
\item We demonstrate substantial improvements (1.4-1.8$\times$) in cross-modal retrieval across six bidirectional tasks, with ablation studies confirming each component's necessity.
\end{itemize}

\begin{figure*}[t]
    \centering
    \includegraphics[width=0.9\textwidth]{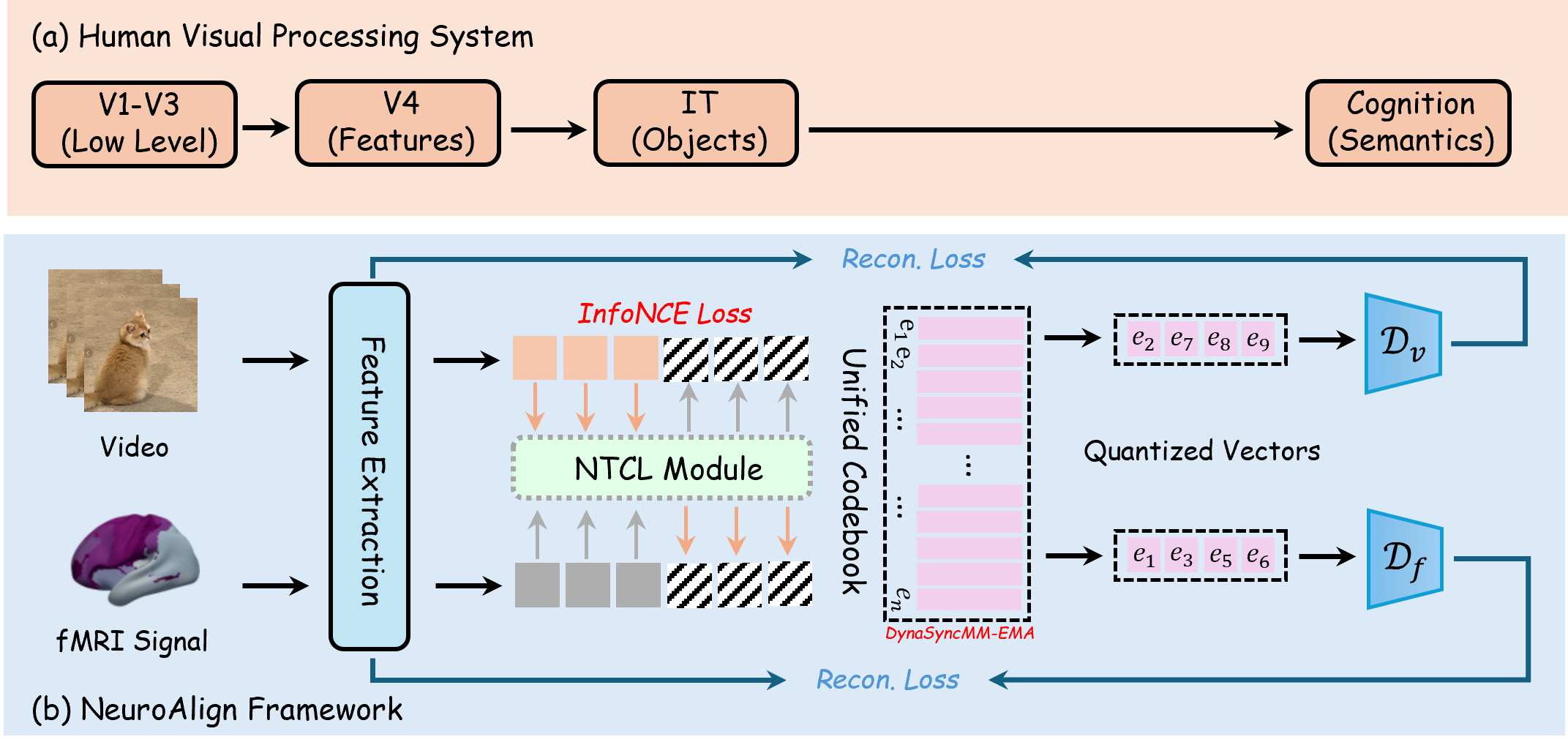}
    \caption{Overview of NeuroAlign framework architecture. The illustration demonstrates our approach using {video, fMRI} data pairs, highlighting the Neural-Temporal Contrastive Learning (NTCL) module and Vector Quantization with DynaSyncMM-EMA update mechanism. While shown with two modalities for clarity, our framework naturally extends to tri-modal integration, enabling comprehensive alignment across $\{$fMRI, Video, Caption$\}$}
    \label{fig:framework}

\end{figure*}

\section{Methodology}
\label{sec:method}

\subsection{Framework Overview}

Our framework addresses fMRI-video alignment through three stages: modality-specific encoders extract features, features are transformed into a shared semantic space, and vector quantization discretizes continuous features into discrete codes: $\mathbf{z}_i = f_{vq}(f_s(f_e(\mathbf{x}_i^m)))$, where $m \in \{f,v,t\}$. This mirrors how the visual cortex progressively abstracts information from low-level features to high-level concepts.

\subsection{Neural-Temporal Contrastive Learning}

Standard contrastive learning treats each sample independently. This fails for fMRI-video alignment because brain responses unfold over time with delays and temporal dependencies. We need a mechanism that explicitly captures temporal dynamics.

Neural-Temporal Contrastive Learning (NTCL) models temporal relationships in both fMRI and video modalities. Given paired sequences $(\mathbf{x}_i^f, \mathbf{x}_i^v)$, we first extract temporal features through encoders. For fMRI, we apply temporal adaptation to compensate for HRF delays: $\mathbf{h}_i^{f'} = \sum_{t=1}^T \alpha_t \mathbf{W}_t\mathbf{h}_i^f[t]$, where learnable weights $\alpha_t$ adjust for the 6-10 second lag between stimulus and peak response.

We then predict future video features from past fMRI context: $\hat{\mathbf{h}}_{i,t+k}^v = g_v(c_f(\mathbf{h}_{i,\leq t}^{f'}))$, where $c_f$ is a bidirectional transformer that captures temporal context. This prediction task forces the model to learn semantically meaningful alignments that respect temporal structure. The training objective uses InfoNCE loss:
\begin{equation}
\mathcal{L}_{ntcl} = -\mathbb{E}\left[\log \frac{\exp(\tau^{-1}s(\hat{\mathbf{h}}_{i,t+k}^v, \mathbf{h}_{i,t+k}^v))}{\sum_{j} \exp(\tau^{-1}s(\hat{\mathbf{h}}_{i,t+k}^v, \mathbf{h}_{j,t+k}^v))}\right]
\end{equation}
where $s(\cdot,\cdot)$ is cosine similarity and $\tau=0.07$ is temperature. This establishes global semantic alignment while accounting for temporal dynamics.

\subsection{Fine-grained Pattern Matching}

NTCL provides global alignment, but fMRI signals are noisy and hierarchical. Continuous embeddings are unstable under physiological noise and cannot capture the discrete, compositional nature of neural coding. We need representations that are both noise-robust and hierarchically structured.

We use Vector Quantization (VQ-VAE) \cite{van2017neural} to discretize features into a learned codebook. Each continuous feature is mapped to its nearest codebook entry: $\mathbf{z}_i^m = \mathbf{e}_k$, where $k = \arg\min_j \|\mathbf{h}_i^m - \mathbf{e}_j\|_2$ and $m \in \{f, v, t\}$. Discretization filters noise by forcing features into stable, interpretable codes.

To handle hierarchical processing, we aggregate information across multiple scales:\\ $\tilde{\mathbf{h}}_i^f = \sum_{l=1}^L \alpha_l\text{TempAdapt}_l(\mathbf{h}_i^f, T')$, where $L$ corresponds to different cortical hierarchy levels (V1, V2, V3, etc.) and $\alpha_l$ are learnable weights. Each $\text{TemporalAdapt}_l$ uses dilated causal convolutions with varying dilation rates, mirroring how different visual cortical areas integrate information over different time scales.

The matching objective combines two components. First, temporal consistency accounts for hemodynamic dynamics:
\begin{equation}
\mathcal{L}_{temporal} = |\mathbf{z}_t^m - \text{HRF}(\mathbf{z}_{t-1}^m)|_2^2
\end{equation}
where $\text{HRF}(\cdot)$ models the BOLD response using a canonical double-gamma function. Second, structural alignment preserves multi-scale pattern relationships:
\begin{equation}
\mathcal{L}_{structure} = \sum_{i,j} |\mathbf{D}_l(\mathbf{z}_i^f, \mathbf{z}_j^f) - \mathbf{D}_l(\mathbf{z}_i^v, \mathbf{z}_j^v)|_2^2
\end{equation}
where $\mathbf{D}_l(\cdot, \cdot)$ computes cosine similarity at scale $l$, ensuring that relationships between fMRI patterns mirror those between corresponding visual patterns. The combined objective is:
\begin{equation}
\mathcal{L}_{match} = \mathcal{L}_{temporal} + \beta\mathcal{L}_{structure}
\end{equation}
with $\beta=0.5$ balancing temporal and structural constraints.

\subsection{Dynamic Synchronized Multi-Modal EMA}

Standard VQ codebook updates treat all modalities equally. This is problematic because fMRI has lower SNR than video/text. Sequential updates \cite{xia2024achieving} introduce bias, later updated modalities dominate the shared codebook. We need balanced updates that account for modality-specific noise levels.

DynaSyncMM-EMA synchronizes codebook updates across modalities with variance-aware weighting: $\beta_{\text{dyn}} = 1.0 + \tanh(\log\frac{\sigma^2_{\text{vt}} + \epsilon}{\sigma^2_{\text{f}} + \epsilon})$, where $\sigma^2_{\text{f}}$ and $\sigma^2_{\text{vt}}$ are fMRI and video+text variances. Higher-variance (higher-information) modalities receive higher weights, preventing noisy fMRI from degrading the codebook.
\\
For fMRI, the sufficient statistics are:
\begin{equation}
\begin{aligned}
N_f &= \beta_{\text{dyn}}\, \sum(E^f),\\
W_f &= (E^{f\top}\beta_{\text{dyn}})\big(\lambda Z^f + \tfrac{1-\lambda}{2}(Z^v+Z^t)\big).
\end{aligned}
\end{equation}
All modalities update simultaneously via exponential moving average:
\begin{equation}
\begin{aligned}
N &\leftarrow \gamma N + (1-\gamma) N_{\text{batch}},\\
W &\leftarrow \gamma W + (1-\gamma) W_{\text{batch}}.
\end{aligned}
\end{equation}
with $\gamma=0.99$. This eliminates sequential bias and enables true multi-modal interaction. The commitment loss balances encoder-codebook alignment:
\begin{equation}
\begin{aligned}
\mathcal{L}_{\text{commit}}^a &= \beta\|\phi^a(\mathbf{x}_i^a) - \operatorname{sg}[\mathbf{e}_i^a]\|^2\\
&\quad+ \tfrac{\beta}{2}\|\phi^a(\mathbf{x}_i^a) - \operatorname{sg}[\mathbf{e}_i^b]\|^2.
\end{aligned}
\end{equation}
The full objective combines all components:
\begin{equation}
\mathcal{L}_{\text{total}} = \alpha_1\mathcal{L}_{\text{ntcl}} + \alpha_2\mathcal{L}_{\text{match}} + \alpha_3\mathcal{L}_{\text{commit}}.
\end{equation}
We use $\alpha_1=0.5$, $\alpha_2=0.3$, and $\alpha_3=0.2$ in the experiment.

\section{Experiments}
\label{sec:experiments}

\subsection{Datasets and Implementation}

We utilized the Human Connectome Project (HCP) \cite{van2013wu} for fMRI pre-training (1,200 subjects, 600,000 segments) and CC2017 \cite{wen2018neural} for alignment tasks (training: 18 clips, 4,320 samples; test: 5 clips, 1,200 samples). Video frames were temporally aligned with fMRI volumes using a 6-second hemodynamic delay shift.

Our framework integrates three backbones: ViT-based encoder pre-trained on HCP for fMRI, ResNet101 for video, and BLIP-2 for text, forming \{fMRI, Frame, Caption\} triplets. Training used batch size 64, learning rate $3 \times 10^{-5}$ with cosine annealing, and settings from \cite{chen2023cinematic}. Experiments ran on a single NVIDIA A100 GPU.

\subsection{Cross-Modal Retrieval Results}

\begin{table}[htb]
\caption{Cross-modal retrieval performance (\%).}
\label{tab:cross_modal_retrieval}
\centering
\setlength{\tabcolsep}{3pt}
\resizebox{\linewidth}{!}{
\renewcommand{\arraystretch}{1.1}
\begin{tabular}{@{}lcccccc@{}}
\toprule
\textbf{Method} & \multicolumn{2}{c}{\textbf{F-V}} & \multicolumn{2}{c}{\textbf{F-T}} & \multicolumn{2}{c}{\textbf{V-F}} \\
\cmidrule(lr){2-3} \cmidrule(lr){4-5} \cmidrule(lr){6-7}
 & R@5 & R@10 & R@5 & R@10 & R@5 & R@10 \\
\midrule
Tri-modal CLIP & 5.24 & 10.74 & 6.41 & 12.83 & 5.25 & 10.58 \\
CLIP4Clip \cite{luo2022clip4clip} & 8.31 & 15.79 & 9.46 & 16.91 & 8.33 & 15.76 \\
Brain-CLIP \cite{huang2023brainclip} & 10.38 & 18.86 & 11.52 & 19.97 & 10.39 & 18.85 \\
BrainVis \cite{williams2022brain} & 15.63 & 26.25 & 15.70 & 27.46 & 15.68 & 26.47 \\
NeuroClips \cite{gong2024neuroclips} & 28.29 & 45.87 & 28.42 & 46.01 & 28.24 & 45.91 \\
\midrule
\textbf{NeuroAlign} & \textbf{50.31} & \textbf{64.64} & \textbf{50.31} & \textbf{64.63} & \textbf{50.18} & \textbf{64.35} \\
\bottomrule
\end{tabular}
}
\end{table}

NeuroAlign achieves 1.4--1.8$\times$ improvements across all retrieval directions (Table~\ref{tab:cross_modal_retrieval}). For F-V retrieval, we achieve 50.31\% (R@5) and 64.64\% (R@10), representing a 10-fold improvement over trimodal CLIP and 1.8$\times$ over NeuroClips. Given an fMRI segment, NeuroAlign correctly identifies the corresponding video within the top-5 candidates in 50\% of test cases, compared to 28\% for the previous best method. The balanced performance across all modality pairs (less than 0.3\% variance) demonstrates a truly unified representational space.

Our approach addresses core fMRI challenges through complementary mechanisms: NTCL compensates for HRF-induced delays (6-10 second lag), enabling stable semantic alignment; VQ-based discretization provides robustness to physiological noise by mapping continuous features to discrete codes; hierarchical matching accommodates multi-scale cortical processing (V1→V2→V3→IT), mirroring biological vision.

\begin{figure}[!htb]
    \centering
    \includegraphics[width=0.47\textwidth]{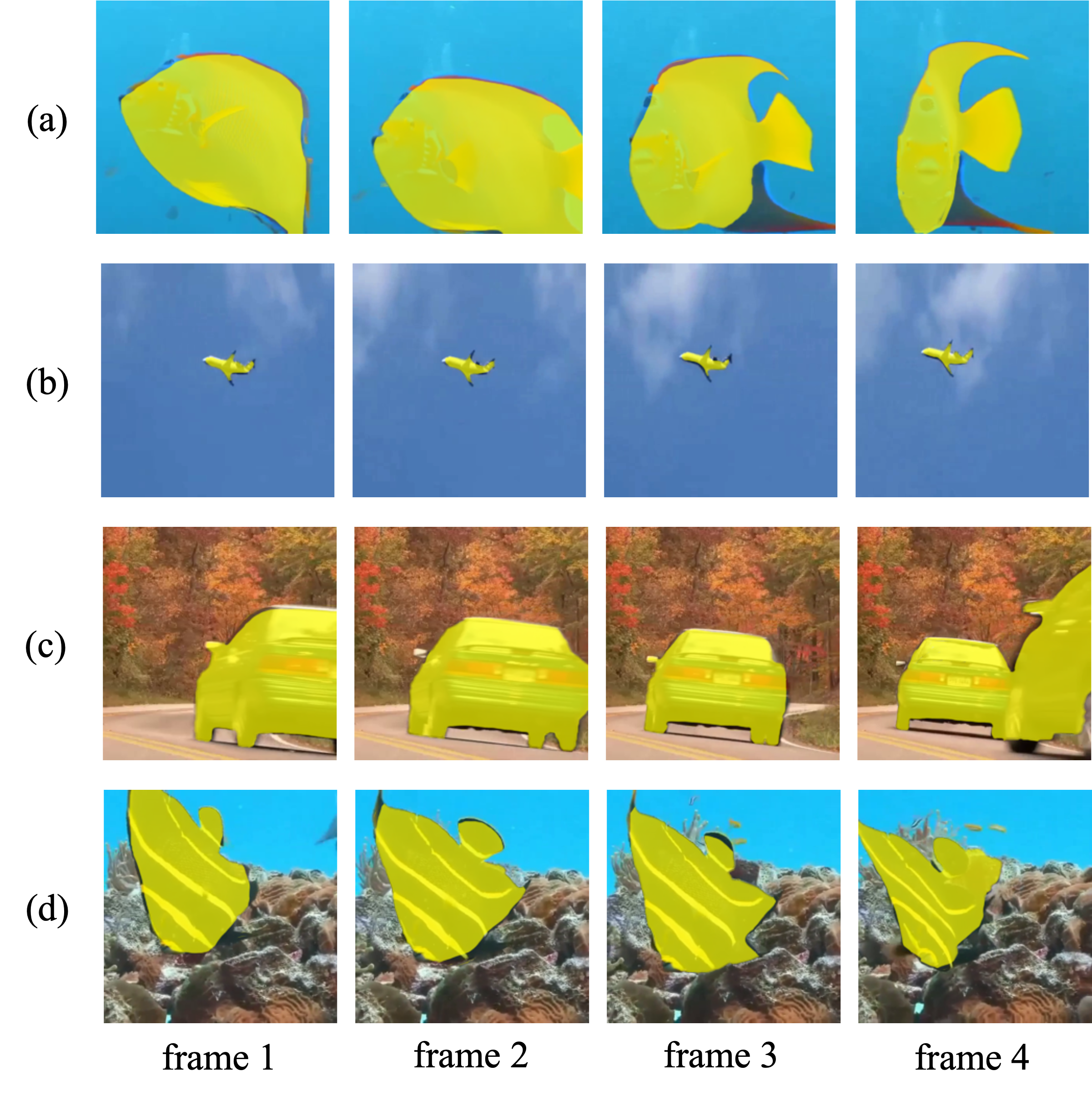}
    \caption{Visualization results of retrieval task. The highlighted areas represent the retrieved video embedding with high activation correspondence with the query fMRI embeddings.}
    \label{fig:highlight}
\end{figure}

\begin{figure}[!htb]
    \centering
    \includegraphics[width=0.47\textwidth]{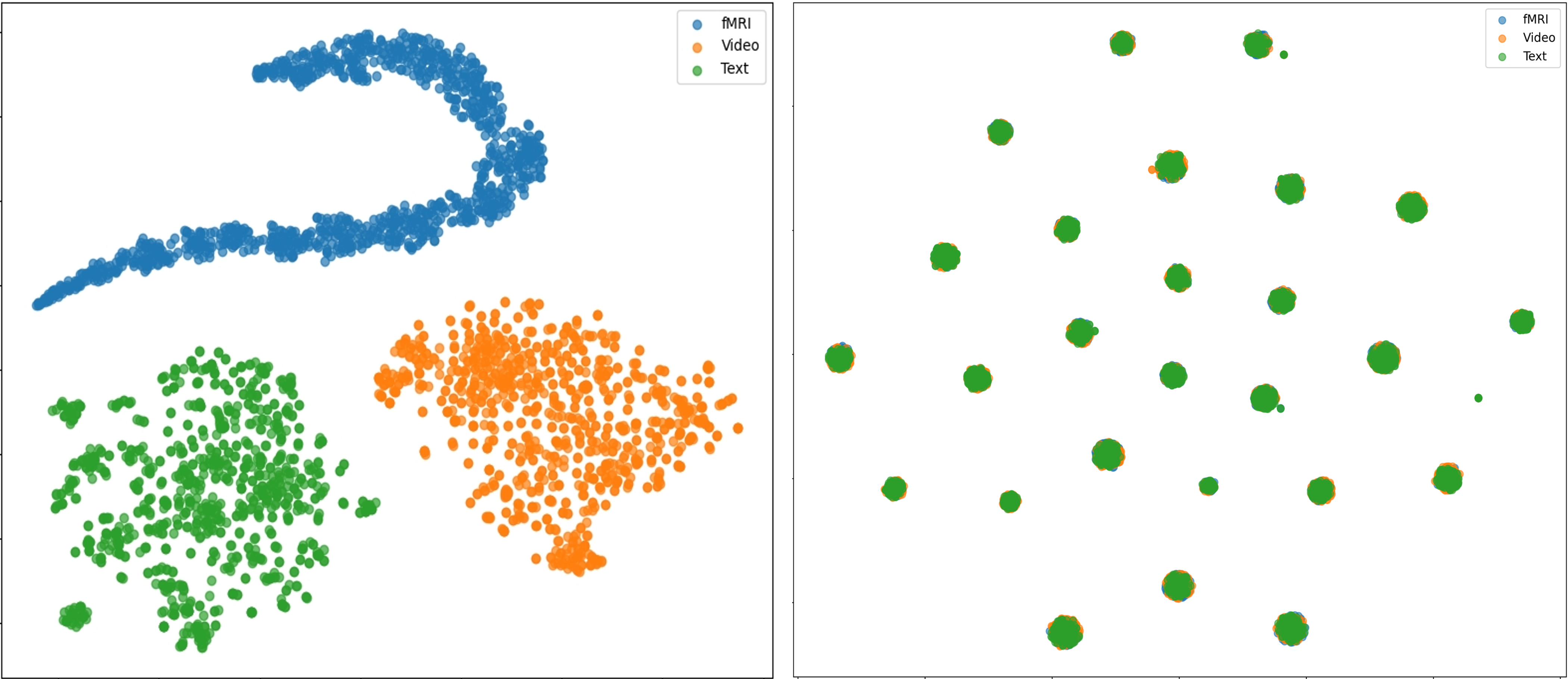}
    \caption{t-SNE visualization comparing NeuroAlign (Ours) with CLIP-based baseline. Colors represent different modalities: fMRI (blue), video (orange), and text (green). NeuroAlign achieves significantly tighter cross-modal clustering compared to CLIP-based methods.}
    \label{fig:ema}
\end{figure}

Qualitative results (Fig.~\ref{fig:highlight}) show that NeuroAlign retrieves correct video frames and identifies spatially localized regions with high activation correspondence to query fMRI patterns. The t-SNE comparison (Fig.~\ref{fig:ema}) reveals superior multi-modal alignment: while CLIP exhibits strong modality-specific clustering, NeuroAlign achieves substantial cross-modal overlap where semantically similar samples from different modalities converge to nearby regions, confirming that discrete VQ-based representations successfully learn shared semantic structure across modalities.

\subsection{Ablation Study}

\begin{table}[htb]
\caption{Ablation study on different components (\%).}
\label{tab:ablation}
\centering
\setlength{\tabcolsep}{3pt}
\resizebox{\linewidth}{!}{
\renewcommand{\arraystretch}{1.1}
\begin{tabular}{@{}lcccccc@{}}
\toprule
\textbf{Method} & \multicolumn{2}{c}{\textbf{F-V}} & \multicolumn{2}{c}{\textbf{F-T}} & \multicolumn{2}{c}{\textbf{V-F}} \\
\cmidrule(lr){2-3} \cmidrule(lr){4-5} \cmidrule(lr){6-7}
 & R@5 & R@10 & R@5 & R@10 & R@5 & R@10 \\
\midrule
w/o Pattern Matching & 20.42 & 30.83 & 20.33 & 30.58 & 20.58 & 31.17 \\
w/o NTCL & 25.49 & 35.88 & 25.63 & 35.93 & 25.44 & 35.93 \\
w/o DynaSync & 35.20 & 48.29 & 35.11 & 48.31 & 35.72 & 48.50 \\
\midrule
\textbf{Full model} & \textbf{50.31} & \textbf{64.64} & \textbf{50.31} & \textbf{64.63} & \textbf{50.18} & \textbf{64.35} \\
\bottomrule
\end{tabular}
}
\end{table}

Table~\ref{tab:ablation} reveals the necessity and complementary roles of each component. Pattern Matching is most critical: its removal causes a 60\% collapse (50.31\%→20.42\% for F-V R@5), validating our core motivation that discrete VQ-based representations are fundamental for stable fine-grained correspondences under fMRI's noise and temporal variability. NTCL removal causes a 49\% drop (50.31\%→25.49\%), demonstrating that temporal modeling and global semantic alignment are essential for compensating HRF-induced delays and establishing cross-modal semantic bridges—the uniform degradation across all tasks (~25\%) confirms NTCL benefits all retrieval directions equally. DynaSyncMM-EMA ablation yields a 30\% reduction (50.31\%→35.20\%), validating synchronized updates with variance-aware weighting to prevent modality bias where later-updated modalities dominate the shared codebook.

Critically, no component can compensate for another's absence, performance gaps between ablated variants and the full model are large and consistent, confirming true synergy rather than redundancy. The full model's 50.31\% R@5 cannot be approached by any two-component subset, demonstrating that fine-grained fMRI-video alignment requires integrating temporal modeling (NTCL), noise-robust discretization (Pattern Matching), and balanced multi-modal fusion (DynaSyncMM-EMA) as stated in our Introduction.

\section{Conclusion}
\label{sec:conclusion}
We presented NeuroAlign, a biologically-inspired framework for fine-grained fMRI-video alignment that mirrors the hierarchical organization of the human visual system. By addressing the fundamental limitations of CLIP-based methods through discrete representation learning, our approach achieves significant improvements (1.4--1.8$\times$) in cross-modal retrieval tasks. Through Neural-Temporal Contrastive Learning that explicitly models temporal dynamics, fine-grained pattern matching via VQ-VAE that provides noise robustness and biological plausibility, and DynaSyncMM-EMA for dynamic multi-modal fusion, NeuroAlign establishes a new paradigm for understanding visual cognitive mechanisms. The framework bridges the gap between computational models and biological vision while providing interpretable insights into neural information processing, with potential applications in clinical diagnostics, neuroprosthetics, and personalized neurotherapies.

\bibliographystyle{IEEEbib}
\bibliography{refer}

\end{document}